\documentclass[runningheads]{llncs}
\usepackage[T1]{fontenc}
\usepackage{graphicx}
\usepackage{tikz}
\usetikzlibrary{shapes.geometric,arrows,fit}
\tikzstyle{process} = [rectangle, 
minimum width=3cm, 
minimum height=1cm, 
text centered, 
text width=3cm, 
draw=black]
\tikzstyle{arrow} = [thick,->,>=stealth]

\begin{document}

\title{Spatial Intelligence of a Self-driving Car and Rule-Based Decision Making}
\author{Stanislav Kikot}
\institute{Sber Automotive Technologies}
%\author{First Author\inst{1}\orcidID{0000-1111-2222-3333} \and
%Second Author\inst{2,3}\orcidID{1111-2222-3333-4444} \and
%Third Author\inst{3}\orcidID{2222--3333-4444-5555}}
%\institute{Princeton University, Princeton NJ 08544, USA \and
%Springer Heidelberg, Tiergartenstr. 17, 69121 Heidelberg, Germany
\maketitle              % typeset the header of the contribution
\begin{abstract}
In this paper we show how rule-based decision making can be combined
with traditional motion planning techniques to achieve human-like 
behavior of a self-driving (SD) vehicle in complex traffic situations.
We give and discuss examples of decision rules in autonomous driving. 
We draw on these examples to illustrate that developing
techniques for spatial awareness of robots  
is an exciting activity which deserves more attention from
spatial reasoning community that it had received so far.

\end{abstract}
\section{Introduction}

In this paper we report on our experience at Sber Automotive Technologies in developing
a motion planner that would be equally fit for urban street, closed areas and 
intercity highways.  
Founded in 2020,  the company possesses a fleet of about 300 autonomous vehicles (AVs) ranging
from golf-cars to semitrailer trucks. 
Out of many aspects of motion-planning here we focus on   
interacting with other agents in a safe, predictable, ethical and confident manner.
We argue that rule-based decision making can play an important role in supplying
robots with this natural to humans and animals `spatial intelligence' ability.

In Section 2 we give a brief description of our SD-system and give a more detailed account of its 
motion planning unit. In Section 3 we highlight a few challenges in designing interaction with other agents 
and discuss their rule-based solutions. In Section 4
we explain how we test our motion planner and rule sets to ensure their quality and avoid regression. The paper ends with a 
brief survey of sources we relied on while creating the planner and a rule-praising conclusion.    

\section{System overview}

\subsection{High-level view of self-driving software}

\begin{figure}
\hbox to \hsize{
\hbox to 8cm{
\vbox to 5cm{
\vfil
\includegraphics[width=8cm]{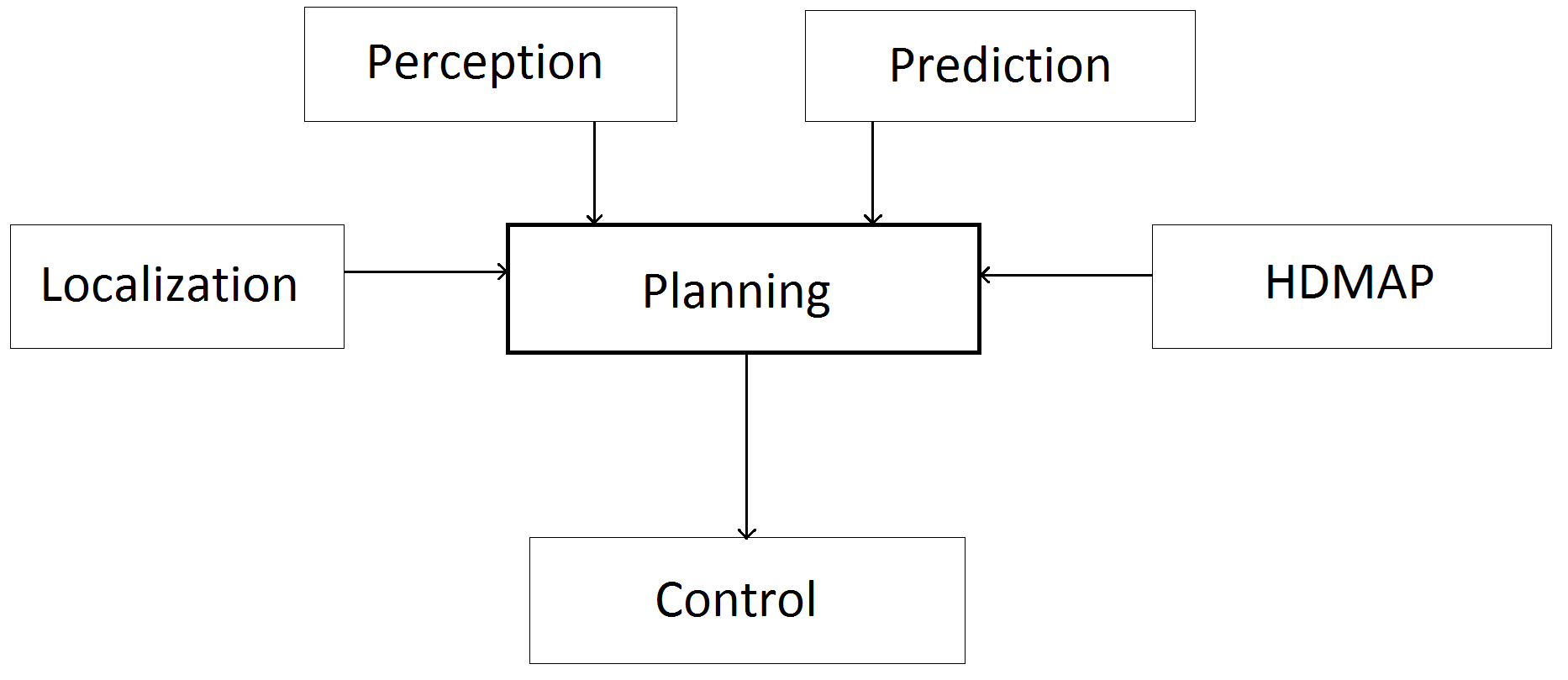}
\vfil
}
}
\hspace*{-3.2cm}
\scalebox{0.75}{
\begin{tikzpicture}[node distance=1.5cm]
\node (lt) [process] {Strategic Planner};
\node (mp) [process,below of=lt] {Maneuver Constructor};
\node (tr) [process,below of=mp] {Shape Planner};
\node (vp) [process,below of=tr] {Velocity Planner};
\node (dm) [process,below of=vp] {Path Selector};
\node (all) [process,fit=(lt)(dm),inner sep=1em] {};
\draw [arrow] (lt) -- (mp);
\draw [arrow] (mp) -- (tr);
\draw [arrow] (tr) -- (vp);
\draw [arrow] (vp) -- (dm);
\draw (-1.95,-2) -- (-7.2,-2.8);
\end{tikzpicture}
}
}
\caption{High-level design of our SD-software and the main parts of its planning unit.} \label{fig0}
\end{figure}
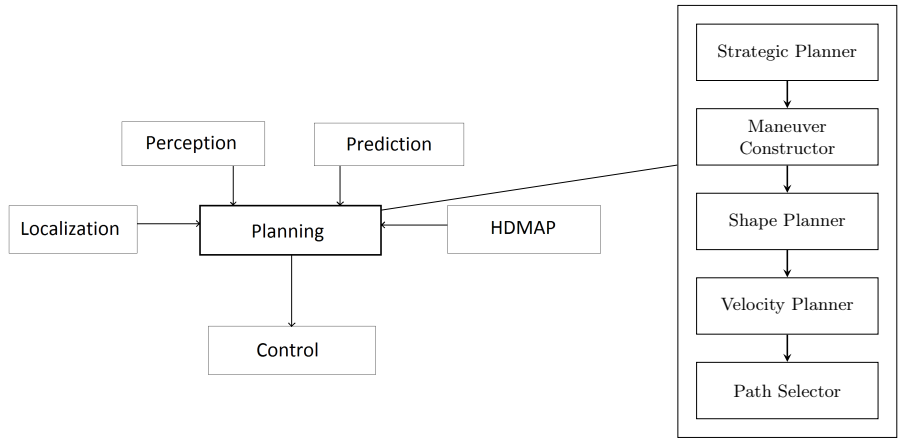

The high-level dataflow of our SD-software is presented on the left of Figure~\ref{fig0}. 
\emph{Perception} integrates information from multiple sensors (such as cameras, lidars and radars),
recognises positions and velocities of other agents as well as static obstacles and traffic light signals 
and creates a `digital representation' of the traffic situation. This representation is 
then passed to \emph{Prediction} which 
supplies each agent with a bunch of possible trajectories and tries to predict
their intentions. All this information is fed to \emph{Planning}. This block 
also has access to
the current position and velocity of the AV coming from \emph{Localization} and
the information about road surface marking and local traffic rules coming from high-definition maps 
of the area. \emph{Planning} guides the AV towards the goal point by 
repeatedly calculating a trajectory endowed with
a velocity profile taking into account the road code and the dynamic environment. 
The \emph{Control} module receives an updated trajectory a few times per second and
drives the AV along it by interacting with steering and velocity actuators through the CAN bus
relying on feedback from \emph{Localization}.

\subsection{Main parts of the motion planner }

\begin{figure}
\includegraphics[width=\textwidth]{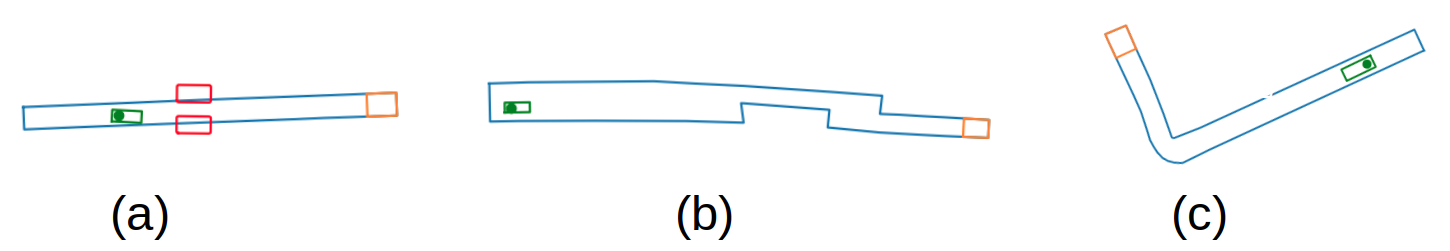}
\caption{Examples of LPTs. In (a) the AV has to pass between two parked cars. Maneuver (b) is obtained
from a `change-left, then change-right' combination in a strategic plan. 
This combination serves to drive around obtacles that block the current lane. The $\sqcap$-shaped cut in (b) results from a solid line on
the road surface between the lanes that the AV should not cross. In (c) there is a bridge support to the right of the AV,
which results in the right road border turning right at right angle. This makes this turn hard to pass.}   \label{fig1}
\end{figure}

Our motion planner has a three-level architecture with decision making at the end. It consists of
the following parts (see Figure~\ref{fig0}, on the right). 
First, a \emph{strategic planner} provides a bunch of strategic plans for reaching the goal point.
A \emph{strategic plan} is a sequence of actions such as `move forward along lane A' 
or `change from lane B into lane C'. Strategic plans are generated using a combination of path-search
techniques in the lane graph of the road network and its sliced version.  
Then \emph{maneuver constructor} turns each of these plans into a \emph{local planning task} (LPT).
Examples of LPTs are  given in Figure~\ref{fig1}.
Each task consists of the initial position of the AV (green rectangle with a bullet inside), 
a target zone (shown in orange), road borders that the AV
should not cross (blue), obstacle areas that must be avoided (red) and penalty zones that
increase the path cost. 
The \emph{shape planner} solves LPTs and produces an optimal curve on the road surface leading from
the initial position of the AV to the target zone avoiding obstacle areas. 
The \emph{velocity planner} takes into account dynamic objects and marks the points of the trajectory
with timestamps when they should be passed (or equivalently, supplies it with a velocity profile).
Sometimes the velocity planner can make a decision to stop at a certain point of the trajectory
(e.g. to give way to a pedestrian).
Finally, the \emph{decision maker} or \emph{path selector} selects one of the calculated trajectories
relying on their comfort and safety metrics.

The reason we put the decision module at the end
is that in some cases such as lane changing in dynamic environment the right decision  
can be made only after the trajectories for two possible courses of action 
(change lane versus push forward) have been fully calculated.

\subsection{Interaction with other agents}

Based on the shape of a candidate path, others' intentions and other aspects of the
traffic situation we single out four ways of interacting with other agents:

\begin{description}
\item{(Ignore)} Ignore the agent
\item{(Drive Around)} React by changing the shape of trajectory
\item{(Give Way)} Stop in front of the agent's route 
\item{(Follow)} Move into the agent's place while keeping distance
\end{description}

For example, we ignore another car if we have right-of-way and the other car moves sufficiently slowly to react and stop.
We give way to a pedestrian if they cross the road, but if a pedestrian walks along the edge of a road, 
we try to overtake it. And if a pedestrian walks away from us in the middle of the road, and we cannot overtake them, then
we follow them. Similarly we follow the vehicles which are in front of us and move away from us, but we give way to 
vehicles that move at angle close to 90 degrees to our trajectory. 

Now a few words about how these reactions are implemented. 
In case of (Ignore) we drive as if the agent did not exist.
For (Drive Around) we generate an obstacle area around the agent's personal space 
and put it into the corresponding LPT. In (Give Way) and (Follow) we react to an agent
by changing not the shape of trajectory but the velocity profile. Thus in these cases no obstacle
areas are added to LPTs and so no steering reaction is produced. Instead we reduce the velocity.
In (Give Way) we stop the AV in front of other agent's personal space. 
In (Follow) we calculate the distance to the agent we must keep for safety and 
plan our velocity accordingly. 

The reason we distinguish between (Give Way) and (Follow) is
that in (Give Way) unlike (Follow) one cannot talk about desirable safety distance
as a function of velocities of the AV and the other agent. 

\section{Instances of rule-based decision making}

\subsection{When to drive around another car?}

A naive rule ``drive around everything that does not move'' may pay off in 
closed areas with low traffic. In multilane urban roads with dense 
traffic it leads to unneeded attempts of lane change in situations when the car
in front of us becomes stationary, but vehicles in adjacent lanes continue 
to move.
An alternative approach is to put on the map all parking slots and 
drive around only those stationary cars that overlap with these areas. 
It may  work in locations when parking rules are strictly enforced
together with a teleoperation service, which allows a teleoperator to
mark particular stationary vehicles as requiring driving around.
We also develop a machine learning solution to this decision problem.

\subsection{Interacting with cars that overtake ego}

By the time we started testing our SD-software on high-speed highways 
we had tested it for a year in urban environment with traffic 
often changing from dense to stationary
with emphasis on safety.  As a consequence, 
that version of the software produced
undesirable braking when other vehicles overtook ego on public highways,
because it tried to stabilize the distance faster than human drivers.

It took a few months to change this behavior
and thouroughly test a new speed controller for safety. During this period our fleet
was supplied with a rule ``ignore cars in front of ego if they
have non-negative acceleration and move sufficiently faster then ego'', which allowed us to test other 
components on highway in parallel with the work on the velocity planner.

\subsection{Pulling away from oncoming cars}

In private roads without median marking it
is important to react to oncoming cars
by pulling away from them to the roadway edge. It took awhile
to figure out the right shape of the corresponding obstacle area.
A naive idea to use the same `personal space area' of other vehicles
as in other cases of interaction, which is based on the bunch of their 
possible trajectories, failed for many reasons.
First, oncoming cars usually, but not always, pull away from us to 
their edge of the road, and it is hard to predict this behavior exactly.
Second, there is a fundamental difference between interacting with
an oncoming car and a static obstacle. If it is not possible to drive
around a static obstacle, the AV should produce no steering reaction, but
this is not so for and oncoming car, where shifting to the edge
of the roadway is always needed. Thirdly, the border of the obstacle area
should be parallel to the road edge, as othwerwise the AV does not
move close to edge sufficiently long. 

\begin{figure}
\includegraphics[width=\textwidth]{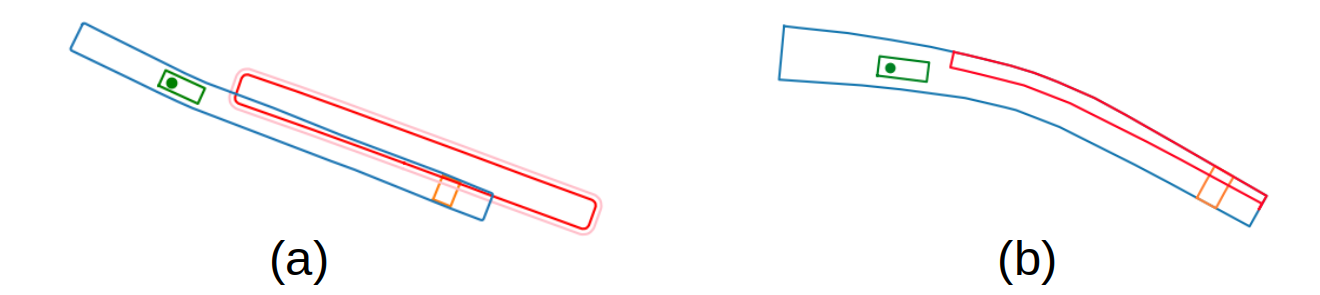}
\caption{Incorrect (a) and correct (b) obstacle areas for oncoming cars.}   \label{fig2}
\end{figure}

These difficulties were overcome by implementing a rule that in presence of 
an oncoming car generates an obstacle area shaped as a narrow strip along
the left border of the driving area. The width of the strip is selected
based on the road geometry and the position of the other car in a 
way to ensure that the resulting LPT has a solution (see Figure~\ref{fig2}). 

\subsection{Ignoring pedestrians that do not intend to get in front of AV}

\begin{figure}
\centerline{\includegraphics[width=11cm]{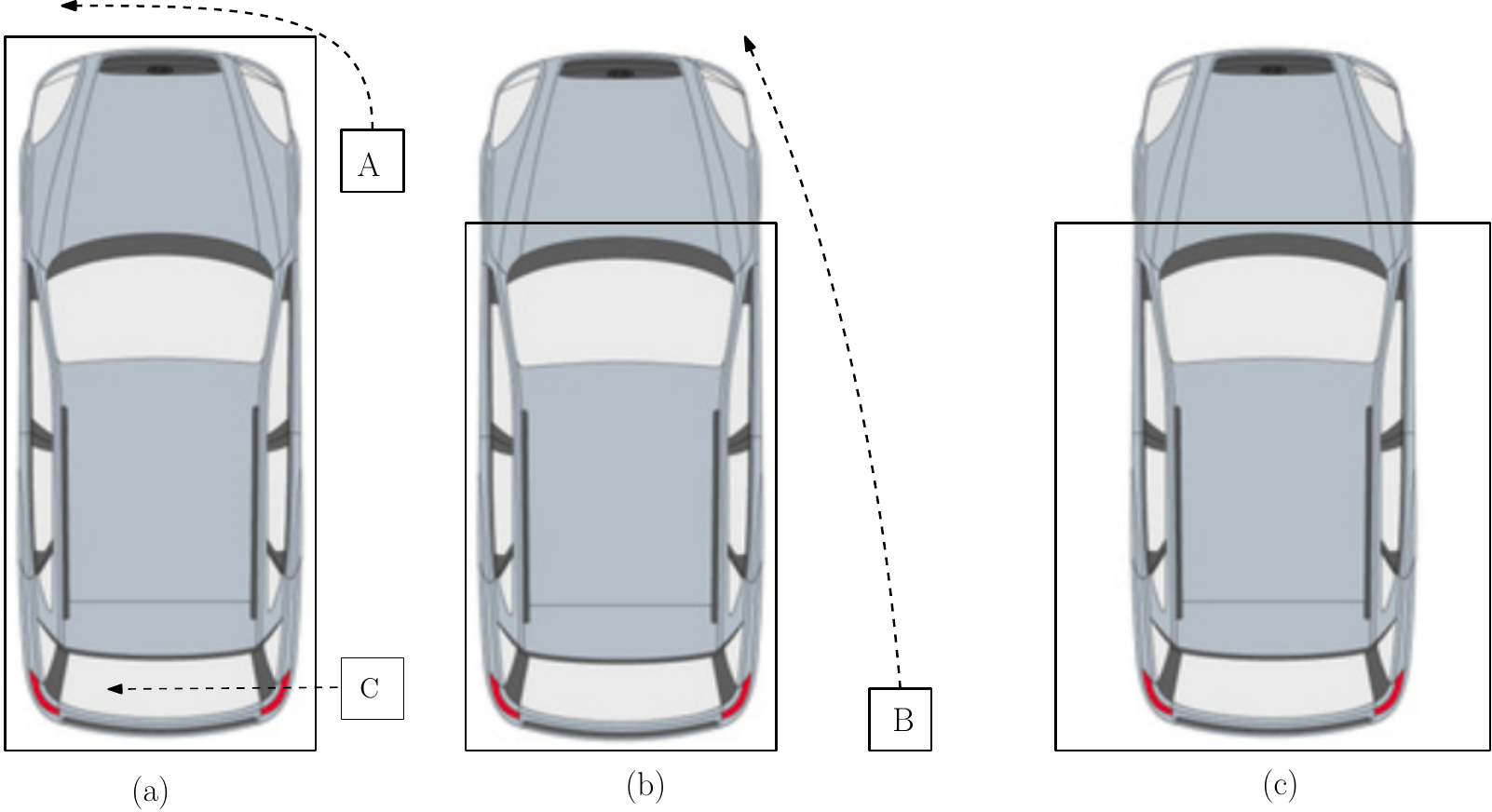}}
\caption{Which pedestrians should be ignored ? (B and C, but not A)}   \label{fig3}
\end{figure}

It is important to distinguish between pedestrians who intend to cross the road
in front of the AV and which require braking reaction  
and those who do not have such intention (such as C in Figure~\ref{fig3}).
To achieve this we introduced a rule ``ignore all pedestrians who collide with the area under the AV''.
First this area was taken as in Figure~\ref{fig3}, (a), then, after studying a case when 
pedestrian A was mistakenly ignored, reduced to (b), and then extended on the sides 
to (c) due to the case with pedestrian B who led to unneeded braking with area (b).

\subsection{Rules versus Finite State Machines and Behavior Trees}

Consider the rule 
``extend by additional 25 cm all obstacles that are further than 12 metres from the AV.''
This rule may seem counterintutive, but at some point we adopted it as a compromise between giving
a hard extension to all objects (which is safer, but may lead to the AV getting stuck 
in places where a human driver can pass) and driving without extra extension of obstacles.
This rule causes the AV to reduce the velocity (up to a stop, if needed) before an obstacle that may 
require the AV to drive around it, and win time to recognise its position, status, 
and shape more accurately and calculate the trajectory around it.
This two-stage behavior 
can be coded using such techniques as finite state machines or behaviour trees. However, we prefer the rule-based approach
due to its succinctness and the fact that you do not need to specially take care of 
state change in case there are many cars parked next to each other.  

\subsection{Yet another controversial rule}

AVs sometimes create dangerous situations by being insufficiently
confident during the `who goes first' negotiations. The rule ``ignore other vehicle, 
if we arrive at the collision point by at least 2 seconds earlier'' can 
make their driving style more assertive and closer to that of an average human driver.
Given that indiscriminate use of this rule may result in dangerous driving, 
we believe that at certain situations this or a similar rule can be adopted. 
Our experiments in the testing ground described in the next section show that it 
improves performance of the AV by 30 cases out of 260 in a scenario, 
where it has to turn left through a dense traffic in the opposite lane. 

\subsection{Give Way or Follow ?}

\begin{figure}
\centerline{\includegraphics[width=6cm]{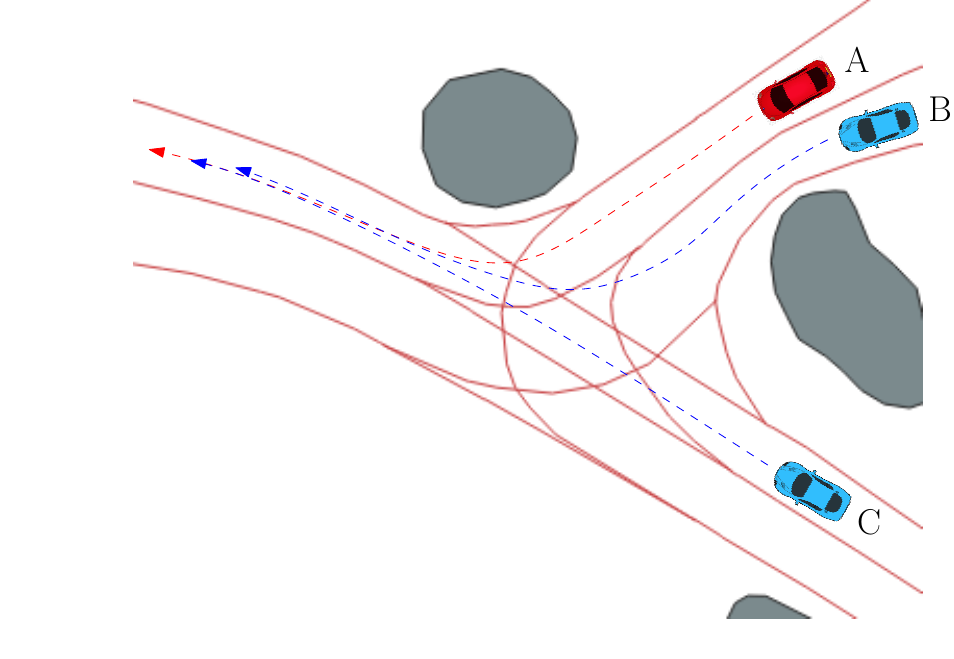}\hspace*{0.5cm}
\includegraphics[width=0.7cm]{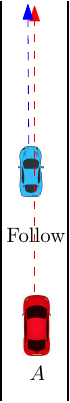}\ \ \ 
\includegraphics[width=2.45cm]{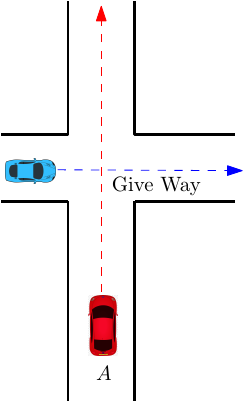}\ \ \
\includegraphics[width=1.71cm]{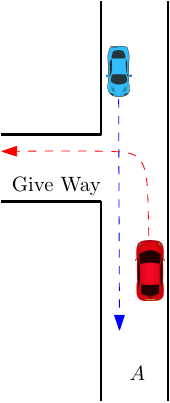}
}
\caption{Car A is the AV. Car B overtakes us, and so requires (Follow) reaction. 
On the other hand car C needs a (Give Way) due to local traffic rules. }   \label{fig4}
\end{figure}

An elegant geometric solution for distinguishing between 
``give way'' and ``follow'' reactions to other cars was discovered
when we faced a junction depicted on the left of Figure~\ref{fig4}.
If the other car is on our path, we compare the angle between our and agent's trajectories with 
55 degrees. Otherwise we do the same and, in addition, 
compare the angle between our and agent's orientations with 45 degrees. It is readily checked 
that under this rule we always `follow' car B, but `give way' to 
car C until it is on our path. This rule also works fine for basic cased on the right of Figure~\ref{fig4}.

\section{Testing and evaluating the motion planner}

Every change in the code of the motion planner undergoes regression testing
in a continuous integration platform, which in addition to unit tests 
runs simulation of the $Prediction+Planning+Control$ part of the SD-pipeline 
and checks that no collision occur in 10 basic scenarios of interaction with other agents.  
In addition, the changes in the shape planner are evaluated on 
a constantly expanding taskset of around 1000 LPTs, where the attention is
paid to the distribution of the time required for task solving, special
quality features of trajectories (such as distance to obstacles 
and road border, smoothness,
accelerations and jerks), and the rate of tasks that are successfully solved.
Finally the change goes into the quality assurance (QA) team, which gives it
an hour-long drive in diverse urban environments and logs 
any noticed defect or change in behavior of the AV. There is also a collection
of a few thousands `night test' simulation scenarios which help to track down 
and correnct rarely occuring regressions.

\begin{figure}
\includegraphics[width=12cm]{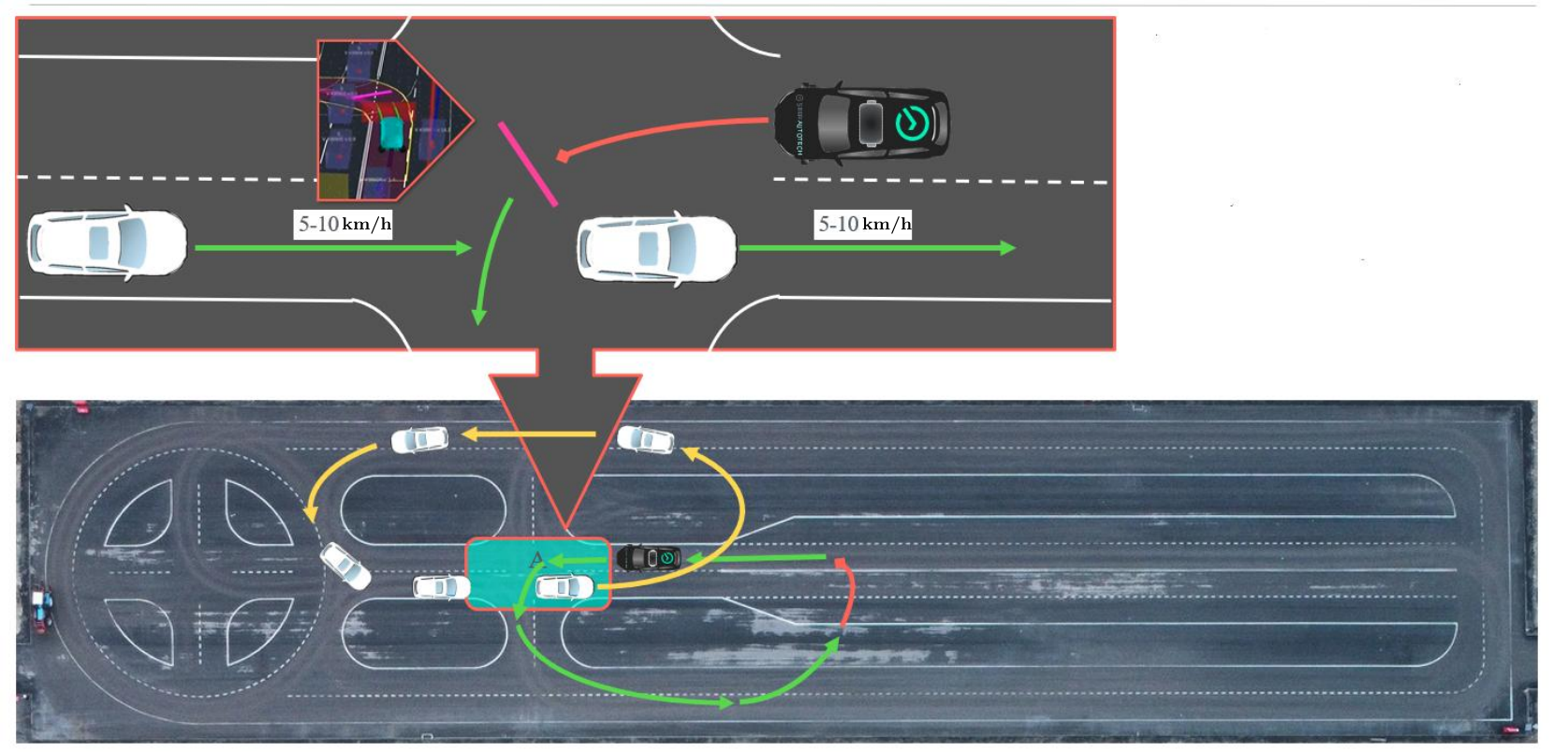}
\caption{Recreation of the `turn left through oncoming traffic' 
scenario in the testing ground. The black car is the AV, the white cars simulate dense oncoming traffic. 
Every fifth white car stops and magnanimously gives way to the AV.}   \label{fig5}
\end{figure}

For A/B testing of different solutions we repeatedly reproduce a
traffic situation in the testing ground and count the numbers of cases
with acceptable and inacceptable behavior of the AV for each of solutions. 
An example of such a simulation is shown in Figure~\ref{fig5}.

\section{Sources of the motion planner}
The strategic planner works on \emph{network of lanelets} introduced in \cite{bender2014lanelets}.
The maneuver constructor, based on Boost Geometry, seems to be original. 
The shape planner uses an optimised version of heuristic search from \cite{botros2021multi} followed
by quadratic and Newtonian spline optimization procedures. The velocity planner runs OSQP solver 
\cite{stellato2020osqp} on the \emph{s-t-graph} of the traffic situation \cite{fan2018baidu}
based on predicted trajectories of other vehicles. Path-velocity decomposition dates back to 
\cite{kant1986toward}.
Our reaction system is inspired by the Mobileye paper on responsibility-sensitive safety 
\cite{shalev2017formal}. However, most of the rules were created by analysing 
cases of incorrect interaction with other agents in urban envoronment reported 
by our QA team.

\section{Conclusion}

The main purpose of this paper is to attract attention of RR community
to practical problems of autonomous driving and demonstrate
that designing logical rules that provide robots with spatial 
intelligence and analysing them integrationally and game-theoretically 
is an exciting activity.
We show that rule-based decision making can play an important
role in SD-software by providing use-cases when logical rules helped to improve
performance of the motion planner. Also rules provide a nice language for
talking about behavior logic of software and analysing bugs: given a case of
undesirable behavior, one may say that this or that rule did or did not work
as expected, and explain why. Finally, some of the rules can be slots and baselines for 
plugging machine learning solutions.

\bibliographystyle{splncs04}
\bibliography{self-driving}

\end{document}